\documentclass[runningheads]{llncs}

\usepackage{eccv}

\usepackage{dsfont}

\usepackage{multirow}
\usepackage{booktabs}
\usepackage{tabularx}
\usepackage{colortbl}
\usepackage{wrapfig,lipsum,booktabs}
\definecolor{cvprblue}{rgb}{0.21,0.49,0.74}
\usepackage{float}
\definecolor{ourpurple}{HTML}{FF7BF6}
\definecolor{ourorange}{HTML}{FF8200}
\definecolor{ourblue}{HTML}{2B83B8}
\definecolor{lightblue}{HTML}{86AAE0}
\definecolor{lightorange}{HTML}{FFC7A7}
\definecolor{lightgreen}{HTML}{BEDFAF}
\definecolor{green}{HTML}{00B341}
\definecolor{lightgray}{HTML}{DBDBDB}

\definecolor{bathcolor}{HTML}{00B1CB}
\definecolor{bedcolor}{HTML}{FFB668}
\definecolor{tablecolor}{HTML}{FF908E}
\definecolor{toiletcolor}{HTML}{9332C9}

\definecolor{lightblue}{RGB}{173, 216, 230} 
\usepackage{eccvabbrv}

\usepackage{graphicx}
\usepackage{booktabs}

\usepackage[accsupp]{axessibility}  

\usepackage{hyperref}

\usepackage{orcidlink}

\begin{document}

\title{Localization and Expansion: A Decoupled Framework for Point Cloud Few-shot Semantic Segmentation} 

\titlerunning{Decoupled Localization and Expansion for PC-FSS}


\author{Zhaoyang Li$^{1\star}$ \orcidlink{0009-0000-6451-5378} \and
Yuan Wang$^{1}$\thanks{Equal 
contribution} \orcidlink{0000-0002-8553-7901}  \and
Wangkai Li$^{1}$ \orcidlink{0009-0008-1776-8450} \and
Rui Sun$^{1}$ \orcidlink{0000-0002-8009-4240} \and
Tianzhu Zhang $^{1,2\,\dagger}$ \orcidlink{0000-0003-1856-9564}
}

\footnotetext[4]{Corresponding author}

\authorrunning{Z. Li et al.}

\institute{MoE Key Laboratory of Brain-inspired Intelligent Perception and Cognition, University of Science and Technology of China \and
Deep Space Exploration Laboratory\\
\email{\{lizhaoyang,wy2016,lwklwk,issunrui\}@mail.ustc.edu.cn, tzzhang@ustc.edu.cn}
}


\maketitle

\begin{abstract}
Point cloud few-shot semantic segmentation (PC-FSS) aims to segment targets of novel categories in a given query point cloud with only a few annotated support samples. 
The current top-performing prototypical learning methods employ prototypes originating from support samples to direct the classification of query points.
However, the inherent fragility of point-level matching and the prevalent intra-class diversity pose great challenges to this cross-instance matching paradigm, leading to erroneous background activations or incomplete target excavation.
In this work, we propose a simple yet effective framework in the spirit of Decoupled Localization and Expansion (DLE).
The proposed DLE, including a structural localization module (SLM) and a self-expansion module (SEM), enjoys several merits.
First, structural information is injected into the matching process through the agent-level correlation in SLM, and the confident target region can thus be precisely located.
Second, more reliable intra-object similarity is harnessed in SEM to derive the complete target, and the conservative expansion strategy is introduced to reasonably constrain the expansion.
Extensive experiments on two challenging benchmarks under different settings demonstrate that DLE outperforms previous state-of-the-art approaches by large margins.
\keywords{Point Cloud Analysis \and Few-shot Learning \and Semantic Segmentation}
\end{abstract}

\section{Introduction}
\label{sec:intro}
As a fundamental computer vision task that boasts broad applications across autonomous driving, robotics, and beyond, point cloud semantic segmentation has propelled conspicuous achievements attributed to elaborate algorithms and well-established datasets.
However, the capabilities of fully supervised segmentation models are restricted to predefined training categories, forming a close-set paradigm that severely limits their practical applications.
In pursuit of human-like intelligence of recognizing novel categories from minimal examples, point cloud few-shot segmentation (PC-FSS)~\cite{zhao2021few} has attracted increasing interest recently, which facilitates the segmentation of new category point clouds (query points) with only a few labeled point clouds (support points) without laborious model retraining.

\begin{figure*}[t]
	\centering
	\includegraphics[width=1.0\linewidth]{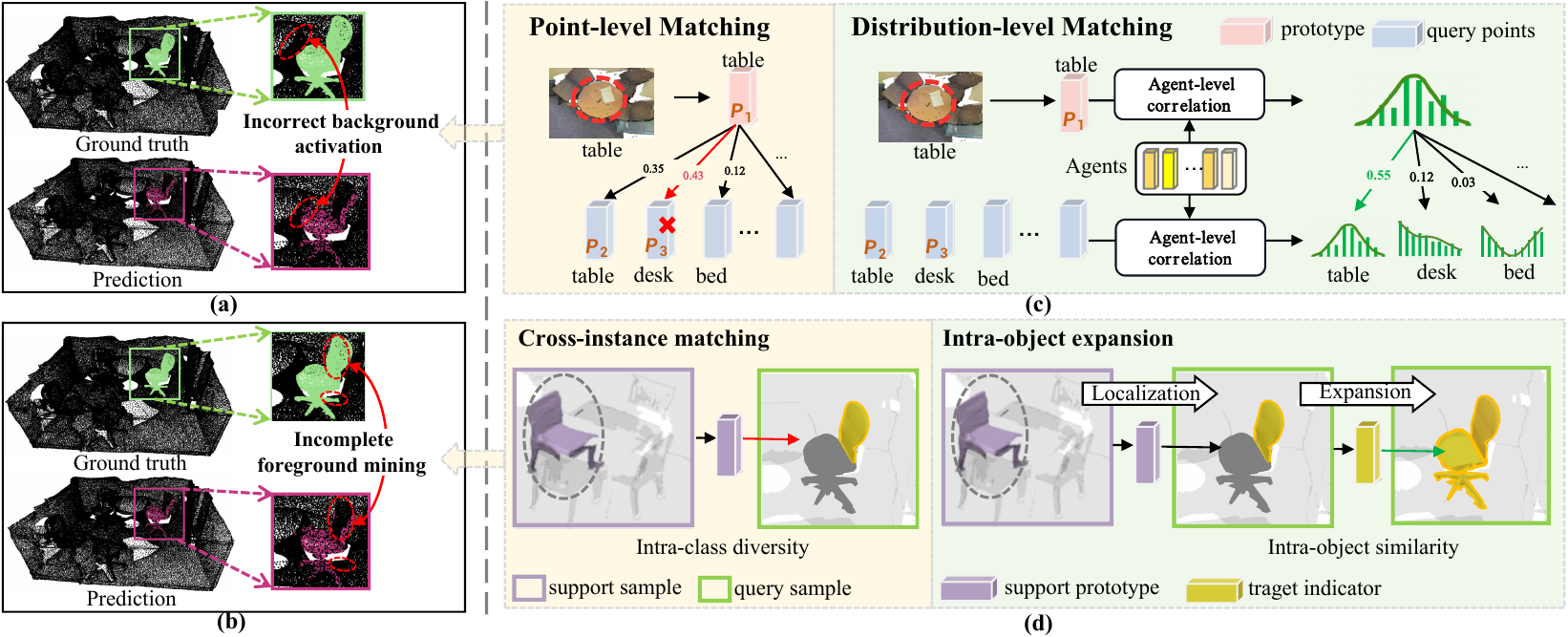}
	\caption{Motivation of our method. (a), (b) Different types of segmentation deficiencies resulted from the inherent fragility of point-level matching and the prevalent intra-class diversity. (c) We employ distribution-level matching that incorporates structural information to replace point-level matching. (d) We leverage intra-object similarity to fully excavate targets, thereby circumventing the impact of intra-class diversity.
 }
	\label{fig:fig1}
\end{figure*}

Prototypical learning~\cite{snell2017prototypical} has emerged as the mainstream paradigm of current top-performing PC-FSS methods~\cite{zhao2021point,ning2023boosting}. Specifically, these approaches condense the support features into a set of representative prototypes, which are then utilized to guide the classification of the query points.
Though achieving promising progress, the significant scene discrepancies result in the direct matching between query points and support prototypes struggling to excavate the target, culminating in two primary types of deficiencies:
\textbf{(1) Incorrect background activation.} The point-level matching primarily focuses on superficial and local consistencies, failing to integrate a broader consideration of semantic structural similarities. The cluttered background renders foreground prototypes susceptible to matching with background points that share local similarities as shown in Figure~\ref{fig:fig1} (a).
\textbf{(2) Incomplete foreground mining.} As illustrated in Figure~\ref{fig:fig1} (b), pronounced intra-class diversities, such as the varieties of scale, pose, and so on, are common occurrences between the support and query objects. This cross-instance variation poses great challenges for support prototypes to furnish query target excavation with essential cues.
To make matters worse, the mistake is inevitably amplified by inbuilt low-data regimes of PC-FSS, leading to sub-optimal results.

Some works~\cite{ning2023boosting, zhang2023few,zhang2023prototype} seek to mitigate this issue by equipping support prototypes with query information via the attention mechanism. Though the query-awareness of support prototypes can be improved in specific scenarios, the reliance on cross-instance prototypes and the fragile nature of point-level matching render these approaches highly sensitive to scene composition.

 \begin{wraptable}{r}{0.4\textwidth}
\renewcommand{\arraystretch}{1.2}
    \centering
    \caption{Quantitative measurement of intra- and inter-object similarity.}
    \scalebox{0.6}{
    			\begin{tabular}{cc|cc}
			\bottomrule
				\bottomrule
                \multicolumn{2}{c|}{S3DIS} & \multicolumn{2}{c}{ScanNet} \\
                Inter-object & Intra-object & Inter-object & Intra-object \\
				\hline
				0.25 & \textbf{0.41} & 0.14 & \textbf{0.39} \\
				\bottomrule
		\end{tabular}
    }
    \label{tab:inter-intra}	
\end{wraptable}

\begin{wrapfigure}{r}{0.5\textwidth}  
	\centering
\includegraphics[width=\linewidth]{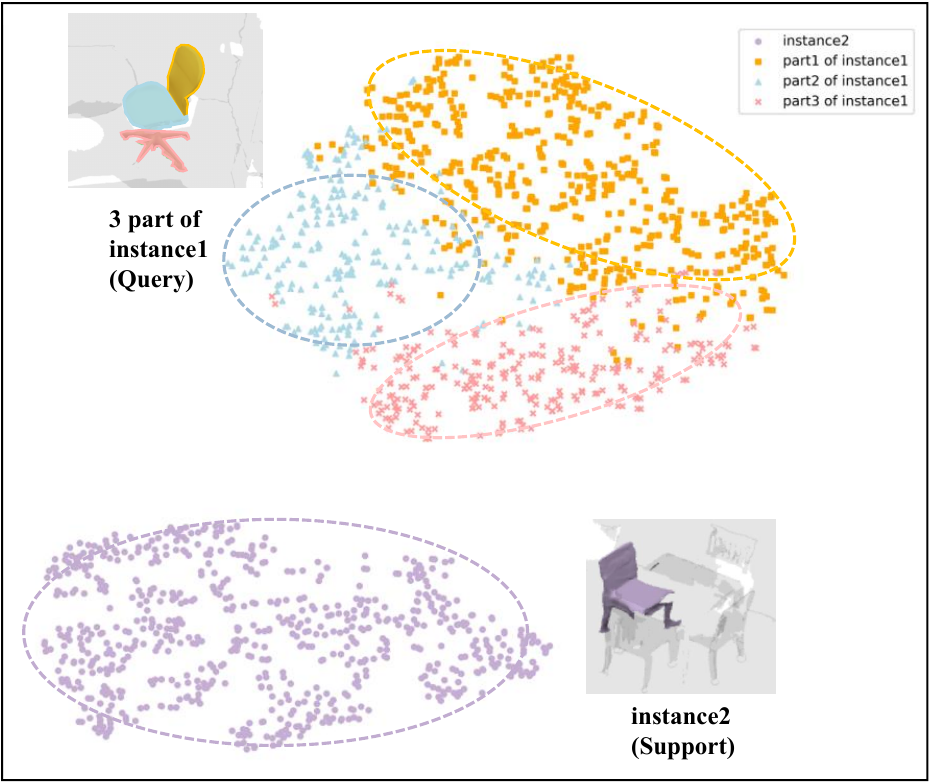}
	\caption{
 T-SNE visualization highlights intra-class diversity and intra-object similarity. 
	}\label{fig:tsne}
\end{wrapfigure}

After an in-depth analysis of the above issues, in this paper, we propose decoupling the challenging PC-FSS into two intuitive procedures to coherently tackle the aforementioned different types of deficiencies,\ie, \textbf{localization and expansion}.
On the one hand, building upon the observation that pixels within the same object share higher similarity than those belonging to different objects~\cite{koffka2013principles,fan2022self} (as shown in Figure~\ref{fig:tsne} and Table~\ref{tab:inter-intra}), we deem that partially activated query object features can serve as more reliable cues to infer complete target, as intra-object features are insusceptible to cross-instance variation.
On the other hand, to alleviate the mismatches attributable to direct point-level comparison, we propose to compare the support prototypes and query points from the perspective of correlations on a set of representative reference features (referred to as \textit{agents}). In this way, high-level semantic structural consistency is integrated to form a more comprehensive distribution-level matching paradigm, thereby achieving more precise target localization.

We propose a simple yet effective PC-FSS framework in the spirit of \textbf{D}ecoupled \textbf{L}ocalization and \textbf{E}xpansion (DLE), which optimizes a structural localization module (SLM) and a self-expansion module (SEM).
\textbf{To precisely locate the targets,} we introduce a set of semantic agents to rectify the direct point-level matching process. The main idea is, for each query point or support prototype, we can obtain the agent-level correlation (\ie, a likelihood vector) by comparing it with all the agents. In essence, the agent-level correlation reflects the consensus among representative agents with a broader receptive field, thus implicitly integrating extensive structural information into the similarity measurement. 
Intuitively, the positive pair (\eg, $P_1$ and $P_2$ in Figure~\ref{fig:fig1} (c)) exhibits consistent similarity distribution across the agent sequence, while the prototype-point pairs (\eg, $P_1$ and $P_3$ in Figure~\ref{fig:fig1} (c)) that possess merely local similarities, lacking analogous agent-level correlation, are suppressed. Confident query target regions (albeit potentially incomplete) can thereby be located through the SLM.
\textbf{To completely dig out the target region,} in SEM, the confident point features activated by SLM are collected as the target indicator via average pooling, which is then employed to capture the rest of the object areas based on the intra-object similarity.
Although the target indicator is immune to the impact of intra-class diversity, it still risks erroneously expanding into the background due to potential similar interferences. 
To overcome this issue, we elegantly design a conservative expansion strategy to eliminate points that are likely to be background elements.
Specifically, for each point within the region expanded by the target indicator, we identify its most similar counterpart within the entire query point cloud as illustrated in Figure~\ref{fig:fig1} (d). The points whose closest matches do not fall within the original region are filtered out, ensuring that the expansion remains within the consistent target areas.

The contributions of our method could be summarized as follows:
\begin{itemize}
	\item We present a simple yet efficient PC-FSS framework in the spirit of Decoupled Localization and Expansion (DLE) to coherently tackle the different types of deficiencies existing in the current prototypical learning-based approaches.
    \item We introduce the Structural Localization Module (SLM) and Self-Expansion Module (SEM) for precise target localization and complete target excavation. They are elegantly integrated to alleviate the impact of scene discrepancies.
    \item Extensive experiments on two challenging benchmarks under different (\ie, 1/2 way, 1/5 shot) settings demonstrate that our approach significantly outperforms previous state-of-the-art PC-FSS approaches.
\end{itemize}

\section{Related Work}
\label{sec:related work}
\subsection{3D Point Cloud Semantic Segmentation}
3D point cloud semantic segmentation is a
fundamental computer vision task that aims to achieve point-wise classification of the given 3D point cloud on predefined categories. Benefiting from the advantages of deep learning~\cite{wu2019pointconv,ye20183d}, remarkable progress has been achieved in parsing 3D scenes by recent deep-learning-based approaches~\cite{huang2018recurrent,landrieu2018large,wang2022semaffinet,lai2022stratified,wang2023rethinking,wang2023focus,pan2023adaptive,wangkai2023maunet,li2023enhancing,sun2023alignment,mai2024pay,wang2024image}. PointNet~\cite{qi2017pointnet} emerged as the pioneering end-to-end neural network designed for direct segmentation of raw point clouds, adeptly maintaining the permutation invariance characteristic of the input data. While PointNet offers a straightforward and efficient architecture, it fails to capture the critical local structures inherent in the spatial distribution of neighboring points. DGCNN~\cite{wang2019dynamic} tackles this limitation by introducing the EdgeConv module, which effectively captures local structures.
Despite their promising results, these methods fail to generalize to novel classes in the low-data regime. In this work, we utilize DGCNN as the backbone of our feature extractor to capture local structure features and propose our method to tackle the 3D point cloud semantic segmentation problem in a few-shot setting. 

\subsection{Few-shot 3D Point Cloud Semantic Segmentation} 
Few-shot 3D Point Cloud Semantic Segmentation aims to perform point-wise classification on point clouds of previously unseen categories with only a handful of labeled
point available. Given the limited availability of annotated examples, it is essential for Few-Shot Segmentation(FSS) tasks to fully leverage the categorial insights present in the support data to improve segmentation accuracy in the query set. Zhao et al.~\cite{zhao2021few} pioneered the exploration of few-shot 3D point cloud semantic segmentation by introducing an attention-aware multi-prototype transductive inference approach, a novel strategy specifically tailored for classifying previously unseen categories with minimal annotated data. Originally conceived for few-shot image classification, ProtoNet~\cite{snell2017prototypical} has been innovatively repurposed for 3D point cloud segmentation in the work of ~\cite{zhao2021few}, which utilizes a DGCNN coupled with a linear projection module for feature extraction, producing support and query features that, after global average pooling to derive class prototypes, are matched using cosine similarity to generate predictions. To reduce contextual gaps between support prototypes and query features,  ~\cite{ning2023boosting} proposed PC-FSS method enhances segmentation by adapting support prototypes to the query context and holistically refining them guided by query features. Despite promising results, the significant intra-class diversities between support and query data persist, limiting the performance ceiling of 3D point cloud few-shot semantic segmentation. In this work, we handle the few-shot point cloud semantic segmentation by adopting the paradigm of~\cite{zhao2021few}. We propose a simple yet effective PC-FSS framework in the spirit of \textbf{D}ecoupled \textbf{L}ocalization and \textbf{E}xpansion (DLE), which leverages the internal similarity within query objects and utilizes holistic category-level cues from support data to effectively discern query objects.

\subsection{Few-shot learning} Few-shot learning targets learning a new paradigm for a novel task with few samples. Previous research predominantly falls into two learning schemes: metric learning and meta-learning.
Metric-learning based approaches~\cite{vinyals2016matching,snell2017prototypical,allen2019infinite,li2019finding,doersch2020crosstransformers,wang2022adaptive}, employ a Siamese network~\cite{koch2015siamese} architecture to process pairs of support and query images, to learn a universal metric that assesses their similarity.
Meta-learning based approaches~\cite{finn2017model,bertinetto2018meta,gordon2018meta,rusu2018meta,cai2018memory,wang2023focus,wang2023rethinking,xiong2024appearance} facilitate swift model adaptation by meta-learning optimization algorithms using a limited number of samples. Based on the paradigm of existing few-shot learning methods, our research improves the performance of point cloud few-shot semantic segmentation with a novel and effective design.

\section{Method}

\label{sec:method}
\subsection{Problem Definition}
The objective of few-shot point cloud semantic segmentation is to endow the segmentation model with the capacity for rapid generalization to novel categories with only a limited set of annotated samples. Following previous works, we adopt the widely used episodic paradigm~\cite{vinyals2016matching}. Specifically, considering the training dataset \( \mathcal{D}_{\text{train}} \) and the testing dataset \( \mathcal{D}_{\text{test}} \), which possess non-overlapping target categories (\( \mathcal{C}_{\text{train}} \cap \mathcal{C}_{\text{test}} = \emptyset \)), we extract a series of subtasks from \( \mathcal{D}_{\text{train}} \) to train the model, while evaluation is kept strictly separate, utilizing the testing dataset \( \mathcal{D}_{\text{test}} \). Each subtask, also referred to as an episode, manifests as an \(N\)-way \(K\)-shot segmentation task within point cloud data. And each episode contains a support set denoted as $\mathcal{S}=\{(I_S^{n,k},M_S^{n,k})\}$, where $I$ is point cloud, $M$ is mask, $k\in\{1,\cdots,K\}$, $n\in\{1,\cdots,N\}$, and a query set \( Q = \{(I_Q, M_Q)\} \). During each episode, the model receives inputs comprising \( \{I_S^{n,k}, M_S^{n,k}, I_Q\} \), where \( I_S^{n,k} \) and \( M_S^{n,k} \) are the point cloud and mask of the support set, respectively, and \( I_Q \) represents the query point cloud. The expected model output is the segmentation prediction for \( I_Q \), with \( M_Q \) as the ground truth for validation. 
In the training phase, the model is optimized to infer the segmentation of the query set \( I_Q \) based on the support set \( S \), guided by the ground truth mask \( M_Q \). During testing, the generalization capabilities of the model are assessed using tasks derived from \( D_{\text{test}} \). To simplify the description, we describe our approach in a 1-way 1-shot setting.

\begin{figure*}[t]
	\centering
	\includegraphics[width=1.0\linewidth]{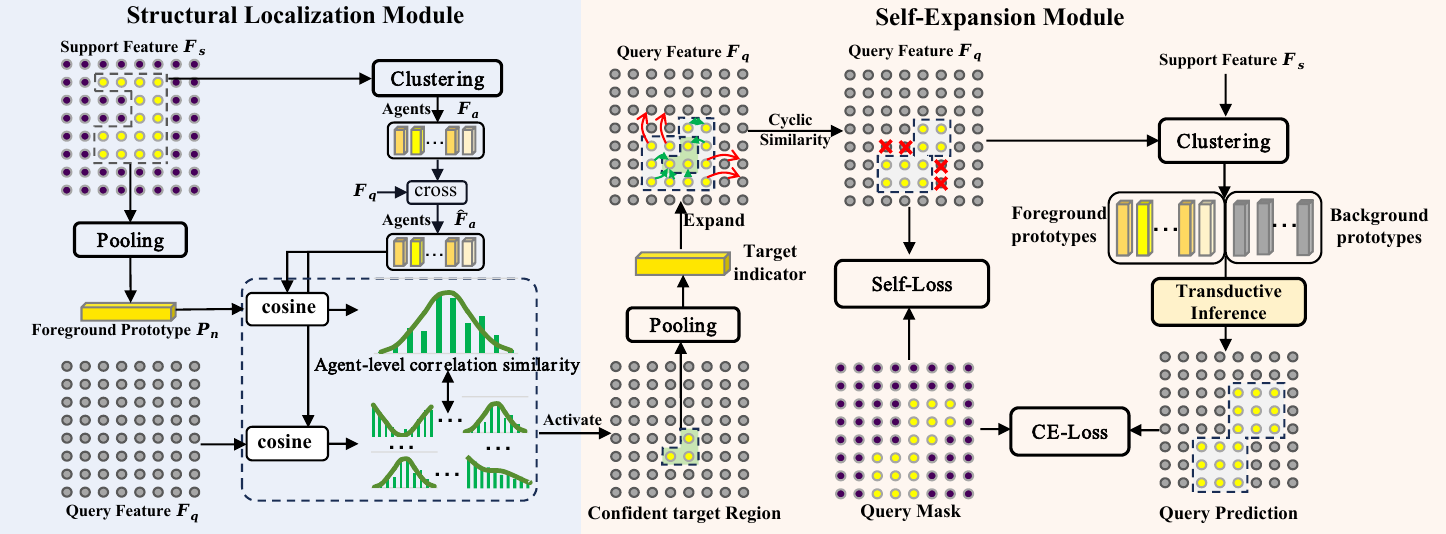}
	\caption{Illustration of the proposed DLE. There are two main modules in DLE. The structural localization module is responsible for precisely locating confident target regions by introducing a set of semantically structure-aware agents. The self-expansion module expands the located target region to mine extensive query information further. These two modules collaboratively constitute a decoupled localization and expansion (DLE) framework for Point Cloud Few-Shot Segmentation.
 }
	\label{fig:framework}
\end{figure*}
\subsection{Overview}
As illustrated in Fig.~\ref{fig:framework}, the proposed framework tailored for Decoupled Localization and Expansion PC-FSS comprises two procedures, i.e., (1) confident target region localization, (2) self-expansion, respectively handled by the structural localization module (SLM) and the self-expansion module (SEM). We locate the most confident target region of the novel class under the support guidance via distribution-level matching in procedure (1). Subsequently, in procedure (2), the self-expansion module expands this initially partial region, enhancing it to more fully capture the target area.
Through procedures (1) and (2), we obtain a pseudo query mask indicating the partial target region of the novel categories. The features of this region are then utilized to derive multiple prototypes, facilitating transductive inference~\cite{zhao2021few}, for producing the final segmentation results. The details are as follows.

\subsection{Structural localization module} 
Considering the significant intra-class diversity between support and query, the direct point-level matching between support prototypes and query points in the procedure (1) for locating the query target can be challenging and susceptible to incorrect matches. To address this issue, we design a structural localization module that innovatively shifts the paradigm from direct point-level matching to a more stringent distribution-level matching through a set of semantically aware agents. Intuitively, agents serve as a pivotal bridge in measuring the similarity between support and query, necessitating a clear perception of semantic information. Therefore, the semantically-aware agents provide a crucial foundation for precise matching in subsequent stages.
To endow agents with semantic awareness, we obtain a set of agents $\mathbf{F}_{a}$ by applying farthest point sampling and K-means clustering on support foreground features, ensuring a diverse and effective setup. Intuitively, the farthest points in feature space can inherently encapsulate diverse aspects of one class (i.e., capturing fine-grained attributes).
Given the initial agents features $\mathbf{F}_{a} = [\mathbf{a}_{1}, \mathbf{a}_{2},..., \mathbf{a}_{N_a}]$ ($\mathbf{a}_{i} \in \mathbb{R}^{1 \times d}$ indicates the $i$-th agent feature), 
we further engage these agents in a cross-attention mechanism with the query features $\mathbf{F}_{q} \in \mathbb{R}^{M \times d}$ (\( M \) signifies the point count and \( d \) the number of the feature channels) to adapt them to the query context. In the cross-attention module, the \emph{queries}~($\mathbf{Q}$) is derived from the agents, while the \emph{keys}~($\mathbf{K}$) and \emph{values}~($\mathbf{V}$) are sourced from the query features $\mathbf{F}_{q}$, formally:
\begin{equation}
	\mathbf{Q}_{i} =
	  \mathbf{a}_{i}\mathbf{W}^{\mathcal{Q}},~ ~
	\mathbf{K}_{j} =
	\mathbf{f}_{j}\mathbf{W}^{\mathcal{K}},~ ~
	\mathbf{V}_{j} =
	\mathbf{f}_{j}\mathbf{W}^{\mathcal{V}},
\end{equation}
among which, $\mathbf{W}^{\mathcal{Q}} \in \mathbb{R}^{d \times d_k}$, $\mathbf{W}^{\mathcal{K}} \in \mathbb{R}^{d \times d_k}$,$\mathbf{W}^{\mathcal{V}} \in \mathbb{R}^{d \times d_v}$ are linear projections.
For the $i$-th $\mathbf{Q}_{i}$, we calculate the attention weights via dot-product between $\mathbf{Q}_{i}$ and all other \emph{keys}:
\begin{equation}\label{eq:attention}
	w_{i, j}=\frac{\exp \left(\beta_{i, j}\right)}{\sum_{j=1}^{M} \exp \left(\beta_{i, j}\right)},
	\beta_{i, j}=\frac{{\mathbf{Q}}_{i} {\mathbf{K}}_{j}^{T}}{\sqrt{d_{k}}},
\end{equation}
where $\sqrt{d_{k}}$ is a scaling factor.
The context-aware agent features are obtained via the weighted sum over all values:
\begin{equation}\label{eq:weight_sum}
	\hat{\mathbf{a}}_{i} =\sum_{j=1}^{M} w_{i, j} {\mathbf{V}}_{j},
\end{equation}
Then a feed-forward network (FFN) is applied to obtain the semantic-aware agents $\hat{\mathbf{F}}_{a} = [\hat{\mathbf{a}}_{1}, \hat{\mathbf{a}}_{2},..., \hat{\mathbf{a}}_{N_a}]$. 
For a given support feature $\mathbf{F}_{s} \in \mathbb{R}^{M \times d}$ and the corresponding mask $\mathbf{M}_{s} \in \mathbb{R}^{M \times 1}$, the prototype for category \( n \) is derived through masked average pooling (MAP) as follows:
\begin{equation}\label{eqn:MAP}
    \mathbf{P}_n = \mathbf{MAP}(\mathbf{F}_s, \mathbf{M}_s).
\end{equation}
Given the semantic-aware agents $\hat{\mathbf{F}}_{a} \in \mathbb{R}^{N_a \times d}$ and the category prototype $\mathbf{P}_{n} \in \mathbb{R}^{1 \times d}$, 
we can define the relation matrix $\mathbf{W_1}$ for the category prototype $\mathbf{P}_{n}$, which encodes the agent-level correlation, formally:
\begin{equation}
	\mathbf{W_1}= \boldsymbol{Sim}(\mathbf{P}_{n}, \hat{\mathbf{F}_{a}}),
\end{equation}
the $\boldsymbol{Sim}$ denotes the cosine similarity.
Similarly, we also define the relation matrix $\mathbf{W_2}$ for the query features $\mathbf{F}_{q}$, which encodes the agent-level correlation, formally:
\begin{equation}
	\mathbf{W_2}= \boldsymbol{Sim}(\mathbf{F}_{q}, \hat{\mathbf{F}}_{a}).
\end{equation}
Given the correlation-encoded matrices $\mathbf{W_1} \in \mathbf{R}^{1 \times N_a}$  and $\mathbf{W_2} \in \mathbf{R}^{M \times N_a}$, a fully connected (FC) layer is initially employed to enhance and integrate features.
Then the correlation matrix  $\mathbf{C} \in \mathbf{R}^{1 \times M}$ is calculated as follows:
\begin{equation}
	\mathbf{C} = \boldsymbol{Sim}(FC(\mathbf{W}_{1}), FC(\mathbf{W}_{2})),
\end{equation} 
\begin{equation}\label{eqn:thread_to_activate}
    \mathbf{M}_{q\tau} = \mathds{1}^{\tau}(\mathbf{C}), 
    \quad
    \mathds{1}^{\tau}(x) =\left\{
	\begin{array}{rcl}
		1, & & x > \tau\\
		0, & & \rm{otherwise}.
	\end{array}\right.
\end{equation}
In N-way settings (\text{\scriptsize N>1}), we obtain N groups of agents $\{\mathbf{F}_{a}^{i}\}_{i=1,2,\ldots,N}$
(\text{\scriptsize $\mathbf{F}_{a}^{i} \in \mathbb{R}^{N_a \times d}$}) from support foreground features, which are then concatenated to form a comprehensive set of agents $\mathbf{F}_{A} \in \mathbb{R}^{(N\times N_a) \times d}$. We then compute the correlation matrix  $\mathbf{C}$ of class prototype $\mathbf{P}_n$ and query features $\mathbf{F}_{q}$ based on their similarity distributions on $\mathbf{F}_{A}$. Subsequently, we employ a comparatively high threshold, denoted as $\tau$, to predominantly identify class-relevant regions while effectively suppressing activation in background areas. In our experimental setup, $\tau$ is set to 0.7. If a point exceeds the threshold for multiple categories, it is assigned to the one with the highest similarity.

\subsection{Self-expansion module}
To further mine extensive query information from the  confident target regions $\mathbf{M}_{q\tau}$ generated by~\cref{eqn:thread_to_activate}, we introduce the self-expansion module (SEM), specifically designed to expand the coverage from localized confident regions to broader target areas by leveraging query intra-object similarity. In specific, utilizing the query features $\mathbf{F}_{q}$ and the confident activation mask $\mathbf{M}_{q\tau}$, we obtain the target prototype $\mathbf{P}_t$ by applying (mask average pooling) MAP to query feature $\mathbf{F}_{q}$, formally:
\begin{equation}\label{eqn:mask_average_pooling}
    \mathbf{P}_t = \mathbf{MAP}(\mathbf{F}_q, \mathbf{M}_{q\tau}).
\end{equation}
Given $\mathbf{F}_{q} \in \mathbb{R}^{M \times d}$ and $\mathbf{P}_t \in \mathbb{R}^{1 \times d}$, we can calculate their similarity matrix as follows:
\begin{equation}
	\mathbf{W_3}= \boldsymbol{Sim}(\mathbf{P}_t, \mathbf{F}_{q}),
\end{equation}
where $\mathbf{W_3} \in \mathbb{R}^{1 \times M}$ indicates every query point's similarity with $\mathbf{P}_t$. Then we can expand the confident region based on the matrix $\mathbf{W_3}$:
\begin{equation}\label{eqn:expand}
    \mathbf{M}_{e\theta} = \mathds{1}^{\theta}(\mathbf{W_3}), 
    \quad
    \mathds{1}^{\theta}(x) =\left\{
	\begin{array}{rcl}
		1, & & x > \theta\\
		0, & & \rm{otherwise}.
	\end{array}\right.
\end{equation}
In our experimental setup, $\theta$ is set to 0.8. If a point exceeds the threshold for multiple categories, it is assigned to the one with the highest similarity.

Although the intra-class similarity within the query is reliable, false matches are still inevitable during the target area expansion process. We adopt a \textbf{\textit{conservative expansion strategy}}, employing a bidirectional confirmation of similarity to reduce false matches further. More precisely, for each point $p$ in $\mathbf{M}_{e\theta}$ that is newly expanded relative to $\mathbf{M}_{q\tau}$, we conduct a cyclic similarity check. This entails identifying the $p'$ in the query that exhibits the highest similarity to $p$. If $p'$ is located within the original confident region $\mathbf{M}_{q\tau}$, it indicates a cyclical consistency between $p$ and points in $ \mathbf{M}_{q\tau}$, thus retaining $p$ as part of the expanded region. Conversely, if $p'$ falls outside $ \mathbf{M}_{q\tau}$, $p$ is likely a misactivated background point and is consequently discarded by setting its corresponding value in $\mathbf{M}_{e\theta}$ to 0.
This process ensures that every newly expanded point in the query maintains cyclic consistency with the initially identified confident activation points. This verification helps prevent the target area's expansion from incorrectly extending into background points, thereby securing a relatively pure target region.

Owing to the inherent clutter in the background, we refrained from applying a similar procedure directly to identify background regions in the query. Here, we obtain $L$ background prototypes $\mathbf{P}_{b} \in \mathbb{R}^{L \times d}$ by employing K-means on support background features, as multiple prototypes enhance the representation of complex and cluttered backgrounds.
But directly utilizing the background prototype $\mathbf{P}_{b}$ to guide the segmentation of query backgrounds can lead to sub-optimal results due to contextual gaps between the support and query. Consequently, we use the mask cross attention (MCA) mechanism that leverages background features within the query to guide $\mathbf{P}_{b}$ in adapting to the query's background. In summary, for background prototype $\mathbf{P}_{b}$, query features $\mathbf{F}_{q}$, and the obtained target mask $\mathbf{M}_{e\theta}$, we perform a mask cross attention operation, akin to Mask2Former~\cite{cheng2022masked}, where \emph{queries}~($\mathbf{Q}$) are derived from $\mathbf{P}_{b}$, \emph{keys}~($\mathbf{K}$) and \emph{values}~($\mathbf{V}$) are sourced from the query features $\mathbf{F}_{q}$, $\mathbf{M}_{e\theta}$ is utilized to execute a masking operation, please refer to the Supplementary Material for more details.
Through the MCA operation, we confine the attention to the query background regions, enabling $P_b$ to adapt more effectively to the background of the query while mitigating interference from foreground points.
To ensure precision in foreground predictions and penalize mispredictions of background points, we devise a self-loss as follows:
\begin{equation}
    {Loss}_{self} = \frac{\sum_{i} (\mathbf{M}_{e\theta}(i) = 1 \land M_{q}(i) = 0)}{\sum_{i} (M_{q}(i) = 0)}.
\end{equation}
where $M_q$ is the query mask, and i indicates coordinate positions. Finally, we follow AttMPTI~\cite{zhao2021few}, integrating a multi-prototype transductive inference approach to derive the ultimate segmentation results.

\newcommand{\pub}[1]{{\color{gray}{\tiny{[{#1}]}}}}
\begin{table}[t]
    \centering
    \tabcolsep 0.05in
    \caption{Results on \textbf{S3DIS} dataset using mean-IoU metric (\%). S$^i$ denotes the split $i$ is used for testing. The best results are shown in \textbf{bold}}
    \label{tbl:exp-s3dis}
    \scalebox{0.7}{
        \begin{tabular}{l|cccccc|cccccc}
            \toprule
            \multirow{3}{*}{\centering \textbf{Method}}& 
            \multicolumn{6}{c|}{\textbf{1-way}} & 
            \multicolumn{6}{c}{\textbf{2-way}} \\
            \cmidrule(lr){2-8} \cmidrule(lr){8-13} 
            & \multicolumn{3}{c|}{\textbf{1-shot}} & \multicolumn{3}{c|}{\textbf{5-shot}} &
              \multicolumn{3}{c|}{\textbf{1-shot}} & \multicolumn{3}{c}{\textbf{5-shot}} \\
            \cmidrule(lr){2-5} \cmidrule(lr){5-8} \cmidrule(lr){8-11} \cmidrule(lr){11-13}
            & S$^0$& S$^1$ & Mean & S$^0$ & S$^1$ & Mean & S$^0$ & S$^1$ & Mean & S$^0$ & S$^1$ & Mean \\
            \midrule
            \midrule
            ProtoNet\pub{NeurIPS'17}~\cite{snell2017prototypical} & 66.18 & 67.05 & 66.62 & 70.63 & 72.46 & 71.55 & 48.39 & 49.98 & 49.19 & 57.34 & 63.22 & 60.28 \\
            MPTI\pub{CVPR'21}~\cite{zhao2021few} & 64.13 & 65.33 & 64.73 & 68.68 & 68.04 & 68.45 & 52.27 & 51.58 & 51.88 & 58.93 & 60.65 & 59.75 \\
            AttMPTI\pub{CVPR'21}~\cite{zhao2021few} & 66.27 & 69.41 & 67.84 & 78.62 & 80.74 & 79.68 & 53.77 & 55.94 & 54.96 & 61.67 & 67.02 & 64.35 \\
            SCAT\pub{AAAI'23}~\cite{zhang2023few} & 69.37 & 70.56 & 69.96 & 70.13 & 71.36 & 70.74 & 54.92& 56.74 & 55.83 & 64.24 & 69.03 & 66.63 \\ 
            ProtoNet+QGE\pub{MM'23}~\cite{ning2023boosting} & 69.39 & 72.33 & 70.84 & 74.07 & 75.34 & 74.71 & 48.98 & 52.62 & 50.8 & 58.85 & 64.26 & 61.56 \\
            AttMPTI+QGE\pub{MM'23}~\cite{ning2023boosting} & 74.30 & 77.62 & 75.96 & 81.86 & 82.39 & 82.13 & 58.85 & 60.29 & 59.57 & 66.56 & \textbf{79.46} & 69.01 \\
            \cellcolor{yellow!12.5}\textbf{Ours} & \cellcolor{yellow!12.5}\textbf{76.54} & \cellcolor{yellow!12.5}\textbf{78.8} & \cellcolor{yellow!12.5}\textbf{77.67} & \cellcolor{yellow!12.5}\textbf{83.15} & \cellcolor{yellow!12.5}\textbf{83.23} & \cellcolor{yellow!12.5}\textbf{83.19} & \cellcolor{yellow!12.5}{\textbf{61.34}}  & \cellcolor{yellow!12.5}{\textbf{63.58}}  & \cellcolor{yellow!12.5}\textbf{62.46} & \cellcolor{yellow!12.5}\textbf{67.92} & \cellcolor{yellow!12.5}74.79 & \cellcolor{yellow!12.5}\textbf{71.36} \\
            \bottomrule
        \end{tabular}
    }
\end{table}
\begin{table*}[t]
    \centering
    \tabcolsep 0.05in
    \caption{Results on \textbf{ScanNet} dataset using mean-IoU metric (\%). S$^i$ denotes the split $i$ is used for testing. The best results are shown in \textbf{bold}}
    \label{tbl:exp-scannet}
    \scalebox{0.7}{
        \begin{tabular}{l|cccccc|cccccc}
            \toprule
            \multirow{3}{*}{\centering \textbf{Method}}& 
            \multicolumn{6}{c|}{\textbf{1-way}} & 
            \multicolumn{6}{c}{\textbf{2-way}} \\
            \cmidrule(lr){2-8} \cmidrule(lr){8-13} 
            & \multicolumn{3}{c|}{\textbf{1-shot}} & \multicolumn{3}{c|}{\textbf{5-shot}} &
              \multicolumn{3}{c|}{\textbf{1-shot}} & \multicolumn{3}{c}{\textbf{5-shot}} \\
            \cmidrule(lr){2-5} \cmidrule(lr){5-8} \cmidrule(lr){8-11} \cmidrule(lr){11-13}
            & S$^0$& S$^1$ & Mean & S$^0$ & S$^1$ & Mean & S$^0$ & S$^1$ & Mean & S$^0$ & S$^1$ & Mean \\
            \midrule
            \midrule
            
            ProtoNet\pub{NeurIPS'17}~\cite{snell2017prototypical} & 55.98 & 57.81 & 56.90 & 59.51 & 63.46 & 61.49 & 30.95 & 33.92 & 32.44 & 42.01 & 45.34 & 43.68 \\
            MPTI\pub{CVPR'21}~\cite{zhao2021few} & 52.13 & 57.59 & 54.86 & 62.13 & 63.73 & 62.93 & 36.14 & 39.27 & 37.71 &43.59 & 46.90 & 45.25 \\
            AttMPTI\pub{CVPR'21}~\cite{zhao2021few} & 56.67 & 59.79 & 58.23 & 66.70 & 70.29 & 68.50 & 40.83 & 42.55 & 41.69 & 50.32 & 54.00 & 52.16 \\
            SCAT\pub{AAAI'23}~\cite{zhang2023few} & 56.49 & 59.22 & 57.85 & 65.19 & 66.82 &66.00 & 45.24 & 45.90 & 45.57 & 55.38 & 57.11 & 56.24 \\
            ProtoNet+QGE\pub{MM'23}~\cite{ning2023boosting} & 57.40 & 59.31 & 58.36 & 60.83 & 66.01 & 63.42 & 37.18 & 39.28 & 38.23 & 44.11 & 47.01 & 45.56 \\
            AttMPTI+QGE\pub{MM'23}~\cite{ning2023boosting} & 59.06 & 61.66 & 60.36 & 66.88 & 72.17 & 69.53 & 43.10 & 46.79 & 44.95 & 51.91 & 57.21 & 54.56 \\
            \cellcolor{yellow!12.5}\textbf{Ours} & \cellcolor{yellow!12.5}\textbf{61.58} & \cellcolor{yellow!12.5}\textbf{63.08} & \cellcolor{yellow!12.5}\textbf{62.33} & \cellcolor{yellow!12.5}\textbf{68.18} & \cellcolor{yellow!12.5}\textbf{72.72} & \cellcolor{yellow!12.5}\textbf{70.45} & \cellcolor{yellow!12.5}{\textbf{46.12}} & \cellcolor{yellow!12.5}{\textbf{49.82}} & \cellcolor{yellow!12.5}\textbf{47.97} & \cellcolor{yellow!12.5}\textbf{54.34} & \cellcolor{yellow!12.5}\textbf{59.79} & \cellcolor{yellow!12.5}\textbf{57.07} \\
            \bottomrule
        \end{tabular}
    }
\end{table*}

\section{Experiments}
\subsection{Datasets and Evaluation Metric}
\textbf{Datasets.} Our experiments are conducted on two benchmark datasets for 3D point clouds: (1) \textbf{S3DIS}\cite{armeni20163d} which comprises 272 point clouds of diversely styled indoor environments including lobbies, hallways, offices, and pantries. This dataset is annotated with 12 semantic classes and an additional background class for clutter. (2) \textbf{ScanNet}\cite{dai2017scannet} which consists of 1,513 point clouds derived from 707 distinct indoor scenes. Annotations within this dataset categorize 20 semantic classes, with an additional class assigned for regions without annotation. Within each dataset, classes are divided into two distinct subsets for cross-validation purposes, specifically $\mathcal{S}^0$ and $\mathcal{S}^1$. These subsets are alternately used as the test classes $\mathcal{C}_{\text{test}}$ and the training classes $\mathcal{C}_{\text{train}}$.  Further details regarding the partitioning process are provided in the appendix. Consistent with the preprocessing strategy of previous work \cite{ning2023boosting}, we partition rooms into 1m x 1m blocks, from which 2048 points are randomly sampled for each block.

\noindent \textbf{Evaluation Metric.} Following previous works~\cite{zhao2021few,ning2023boosting}, Mean Intersection-over-Union (mean-IoU) is adopted as our evaluation metrics for all experiments. The Intersection over Union (IoU) for individual classes is computed as \(\frac{TP}{TP + FN + FP}\), wherein \(TP\), \(FN\), and \(FP\) represent the counts of true positives, false negatives, and false positives, respectively. In the context of few-shot learning, the mean-IoU is derived by averaging the IoU across all testing classes in the unseen category set \(\mathcal{C}_{\text{test}}\).
\subsection{Implementation Details}
Our approach is implemented using the PyTorch framework, and all experiments are conducted on one NVIDIA GeForce RTX 3090 GPU. The feature extraction module is pre-trained on the training set \( \mathcal{D}_{\text{train}} \) for 100 epochs with a batch size of 32, employing the Adam optimizer with a learning rate of 0.001. After pre-training, we train our model using the Adam optimizer, setting the initial learning rate to \(0.0001\) for the feature extractor---initialized with pre-trained weights---and \(0.001\) for the other modules within the architecture, both of them are halved every 5000 iterations. Following~\cite{zhao2021few}, models undergo 40,000 iterations throughout the training process. During each iteration, we randomly sample an episode, applying Gaussian noise and arbitrary z-axis rotations to augment the point clouds in both the support and query sets.
\subsection{Comparison with State-of-the-Art Methods}
\textbf{S3DIS.}
In Table~\ref{tbl:exp-s3dis}, we compare our proposed DLE with the state-of-the-art few-shot point cloud semantic segmentation methods. It can be observed that our DLE outperforms all previous models under all settings. Specifically, our approach achieves 77.67\% and 62.46\% mIoU in the 1-way 1-shot and 2-way 1-shot settings, surpassing the most competitive QGE~\cite{ning2023boosting} by 1.71\% and 2.89\%, respectively. 
Furthermore, the improvements are more pronounced in the 1-shot setting than in the 5-shot due to the scarcity of support information, which amplifies  intra-class diversity issue. 
To address this, our method leverages distribution-level matching and better intra-object similarity within the query, thereby achieving more precise segmentation.

\begin{figure*}[!t]
	\centering
	\includegraphics[width=1.0\linewidth]{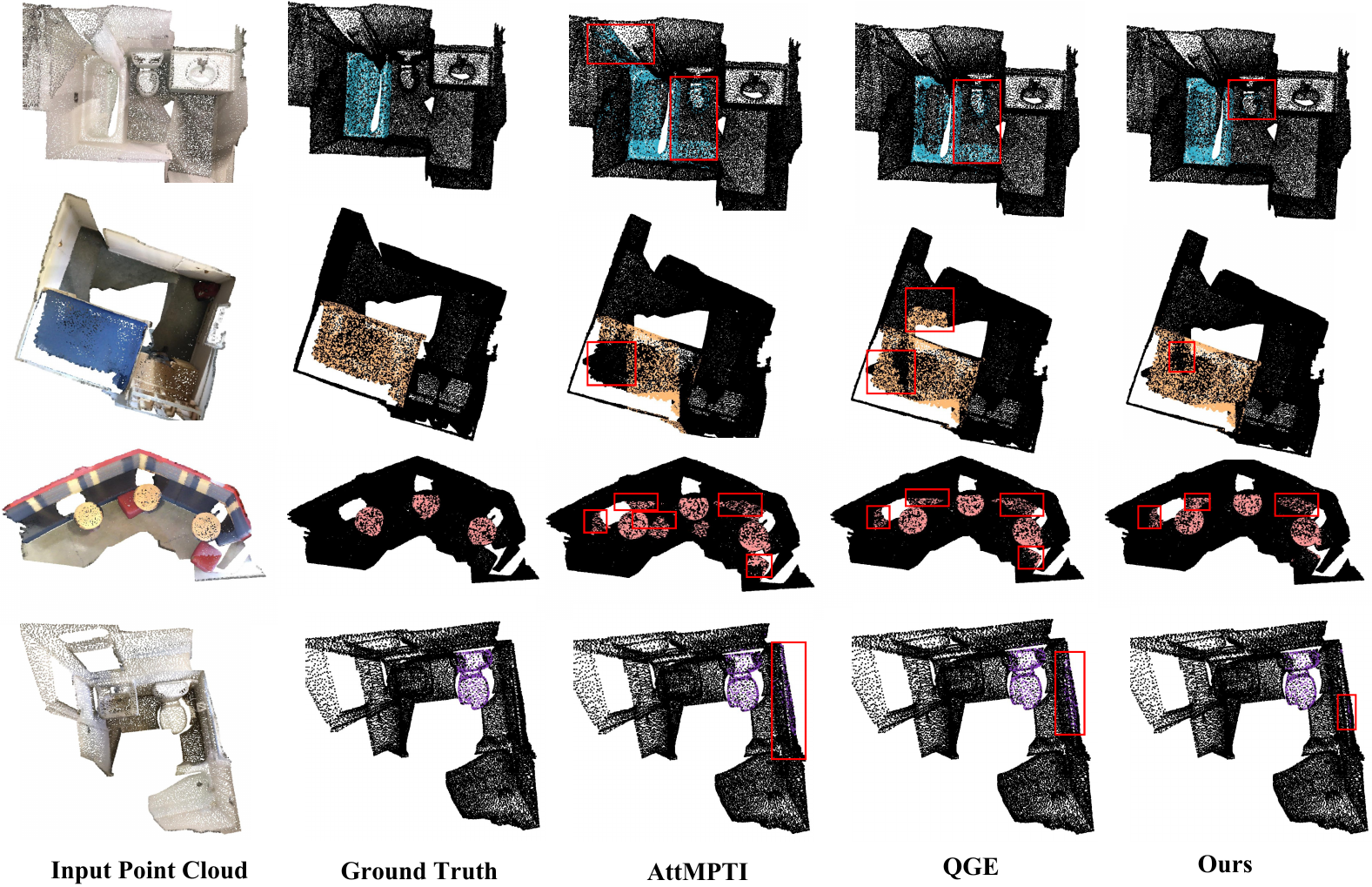}
	\caption{Qualitative results of our method in 1-way 1-shot point cloud few-shot segmentation on ScanNet dataset in comparison to the ground truth, AttMPTI and QGE. The target classes from top to bottom are \textcolor{bathcolor}{"bathtub"} (first row), \textcolor{bedcolor}{"bed"} (second row), \textcolor{tablecolor}{"table"} (third row) and \textcolor{toiletcolor}{"toilet"} (last row).
 }
	\label{fig:results}
\end{figure*}

\noindent \textbf{ScanNet.}
In Table~\ref{tbl:exp-scannet}, we compare results on the more challenging ScanNet dataset, which features diverse room types. Our DLE method still outperforms the best existing method, with gains of 1.97\% and 3.02\% mIoU in the 1-way 1-shot and 2-way 1-shot settings, respectively. Our model shows greater improvements on ScanNet due to its significant intra-class diversity, as shown in Table~\ref{tab:inter-intra}. This aligns with the issues our model addresses, highlighting its robustness in complex environments. 
Additionally, Fig.~\ref{fig:results} shows prediction visualizations, demonstrating the visual superiority of our method.

\subsection{Ablation Study}
A comprehensive series of ablation studies were conducted on the S3DIS $S^0$ split under the 1-way 1-shot setting, employing AttMPTI~\cite{zhao2021few} as our baseline model, to analyze each component of the proposed DLE. 

\begin{wraptable}{r}{0.6\textwidth}
 		 \caption{Ablation study results. Experiments are conducted on S3DIS-$S^0$ for 1-way 1-shot setting.}
 \hspace{2mm}
 \begin{subtable}[h]{0.3\textwidth}
		\centering
		\resizebox{!}{11mm}{
		\begin{tabular}{c|cc|cc}
			\bottomrule
            \multirow{2}{*}{SLM} & \multicolumn{2}{c|}{SEM} &\multirow{2}{*}{mIOU} & \multirow{2}{*}{$\Delta$} \\
			\cline{2-3}
			& Self-loss  & Cycle & \\
			\hline
			\hline
			& & &  66.27 & +0.0 \\      
			
            &  $\checkmark$  &  & 67.9 & +1.63 \\
            & & $\checkmark$&  70.61 & +4.34  \\ 
			& $\checkmark$& $\checkmark$& 71.56 & +5.29 \\
            $\checkmark$&   & & 74.26 & +7.99 \\
			\rowcolor{yellow!12.5} $\checkmark$   & $\checkmark$  &  $\checkmark$  &\textbf{76.54} & +\textbf{10.27}\\
			\bottomrule
	\end{tabular}}
		\caption{Ablation studies of the proposed SLM and SEM modules.}
		\label{tab:ablation}
 		\end{subtable} 
        \hspace{1mm}
 \begin{subtable}[h]{0.25\textwidth}
		\centering
		\resizebox{!}{9mm}{
			\begin{tabular}{c|c}
				\bottomrule
				Init & mIoU \\
				\hline
				\hline
				Random & 72.1 \\
				Learnable & 73.08 \\
                Cluter(all) & 75.56 \\
				\rowcolor{yellow!12.5} Cluster(fore) & \textbf{76.54}\\
				\bottomrule
		\end{tabular}}
		\caption{Ablation studies on different approaches for agents initialization.}
		\label{tab:init}	
 		 \end{subtable}
\end{wraptable}
\noindent \textbf{Effectiveness of the structural localization module.} 
The comparison between the 1$^{st}$ and 5$^{th}$ rows of Table~\ref{tab:ablation}, we can observe a significant performance lift, i.e., 7.99\% in mIoU, demonstrating that our proposed semantically structure-aware agents effectively address the issue of mismatch, thereby accurately identifying discriminative foreground regions in the query point cloud. This provides a solid foundation for subsequent utilization of reliable intra-class similarity within the query to guide its segmentation.

\noindent \textbf{Analysis of Agent Generation.}
To evaluate the effectiveness of different agent generation methods, we compare several intuitive initialization approaches in Table~\ref{tab:init}. The $Random$ method selects $N_a$ point features randomly from the support features. $Learnable$ initializes agents randomly but makes them trainable. $Cluster(all)$ and $Cluster(fore)$ cluster $N_a$ agents from all support points and foreground point features, respectively. The Cluster-based methods consistently outperform others due to their ability to identify meaningful patterns and structures in the data. Specifically, clustering within foreground points is more effective, avoiding background noise and yielding agents that better reflect the structured semantic information of the foreground. This provides a solid foundation for using agents to establish distribution-level matching between support and query.

\begin{wrapfigure}{r}{0.5\textwidth}    
    \centering
	\includegraphics[width=1\linewidth]{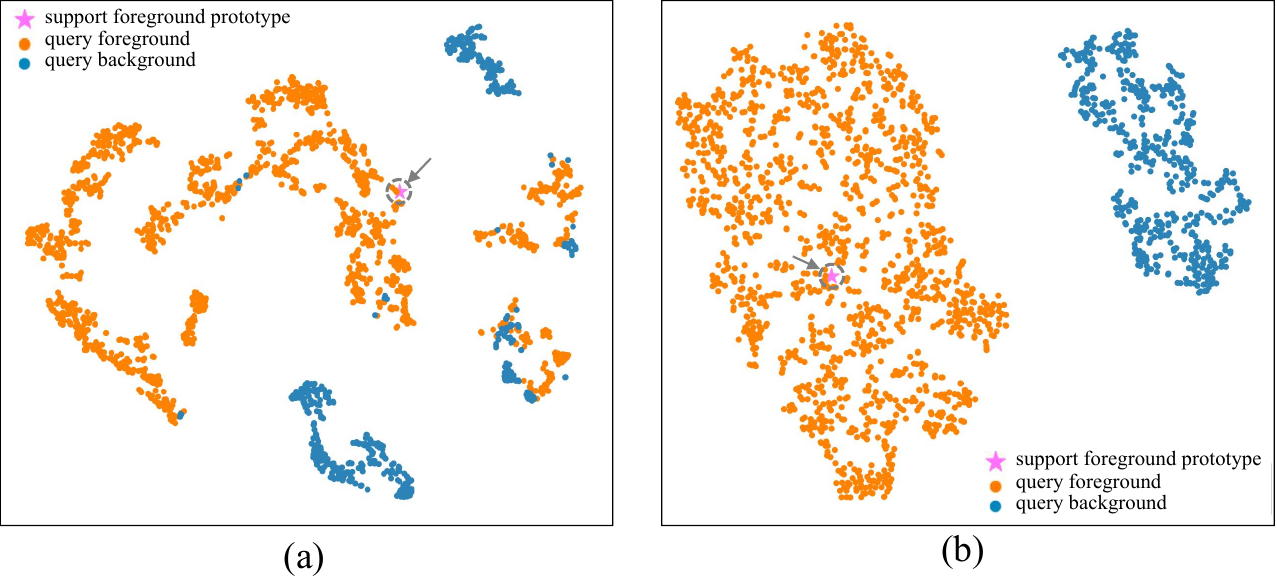}
	\caption{
 T-SNE visualization of the feature distribution of the distribution of \textcolor{ourpurple}{support foreground prototype}, \textcolor{ourorange}{query foreground} and \textcolor{ourblue}{query background} in the original feature space (a) and in the agent-dimension-modified space (b).
 }
    \label{fig:agent_tsne}
\end{wrapfigure}

\noindent \textbf{Investigation of the agent-based distribution-level matching.} To more intuitively demonstrate the effectiveness of our proposed agent-based distribution level matching, we conduct T-SNE visualizations both in the original feature space (Fig.~\ref{fig:agent_tsne} (a)) and in an agent-dimension-modified space (Fig.~\ref{fig:agent_tsne} (b)) for support prototypes and query features. In the agent-dimension-modified space, each point's features reveal its similarity to our semantically aware agents, allowing us to more accurately separate foreground from background with finer detail. This approach not only clarifies the distinction between foreground and background but also aligns the support foreground prototype more closely with the query foreground, enabling a more accurate selection of query foreground.

\begin{wraptable}{r}{0.3\textwidth}
 \centering
 \caption{Ablation studies of the number of agents.}
        \resizebox{!}{5mm}{
        \begin{tabular}{c|cccc}
			\bottomrule
			\bottomrule
			\multicolumn{1}{c|}{N$_a$}  &  \multirow{1}{*}{50} &\multirow{1}{*}{100} &\multirow{1}{*}{150} &\multirow{1}{*}{200} \\
			\hline
            mIoU& 75.79&\textbf{76.54}&76.19&75.06 \\
			\bottomrule
	\end{tabular}}
		\label{tab:agents_num}
\end{wraptable}

\noindent \textbf{Effectiveness of the self-expansion module.} As shown in Table~\ref{tab:ablation}, the full SEM module improves performance by 5.29\% (comparing the 1$^{st}$ and 4$^{th}$ rows). Moreover, the addition of the self-mining module boosts performance by 2.28\% mIoU (comparing the 5$^{th}$ and 6$^{th}$ rows) based on the structured disambiguation localization module. This enhancement is due to SEM's bidirectional confirmation of similarity check and a tailored loss function, which leverages intra-object similarity within the query to excavate more target regions accurately.

\begin{figure*}[!t]
	\centering
	\includegraphics[width=1.0\linewidth]{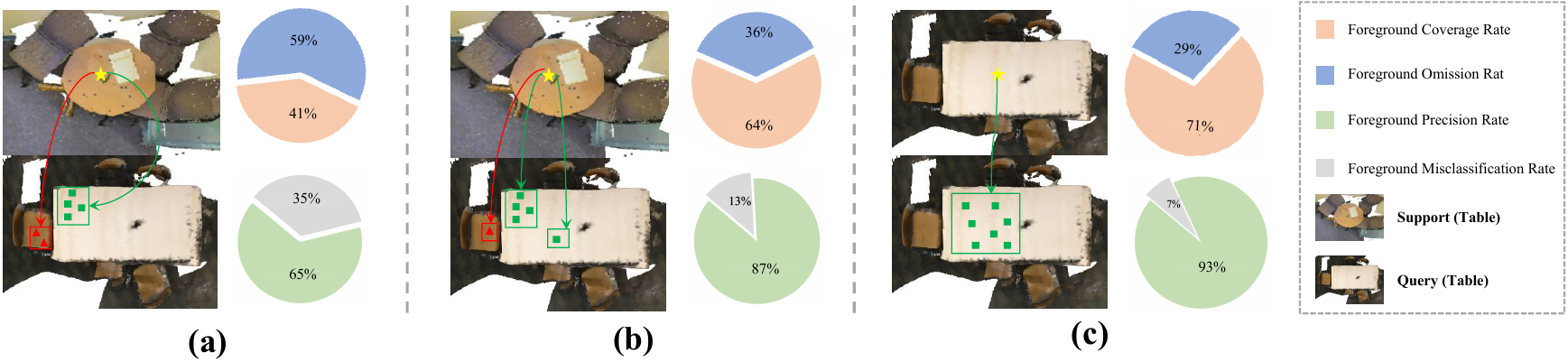}
	\caption{Illustration of point correspondence and statistics data visualization. The red arrows indicate the false matches and the green arrows indicate the correct matches. \textcolor{lightorange}{Foreground Coverage Rate} is defined as the proportion of selected foreground points to all foreground points. \textcolor{lightgreen}{Foreground Precision Rate} is defined as the proportion of correctly identified foreground points among all selected points. 
 }
	\label{fig:pie_chart}

\end{figure*}

\noindent \textbf{Decoupled Localization and Expansion Paradigm Discussion.}
Since FSS models are trained on limited data, the feature space misaligns for new classes, causing poor feature representation and scattered features (intra-class diversities). This hinders direct matching between query points and support prototypes, leading to 1) incorrect background activation and 2) incomplete foreground mining. Figure~\ref{fig:pie_chart} illustrates our solutions: (a) shows that direct matching causes false matches, reducing 
coverage and precision rates for foreground points; 
\begin{wrapfigure}{r}{0.33\textwidth}    
    \centering
	\includegraphics[width=1\linewidth]{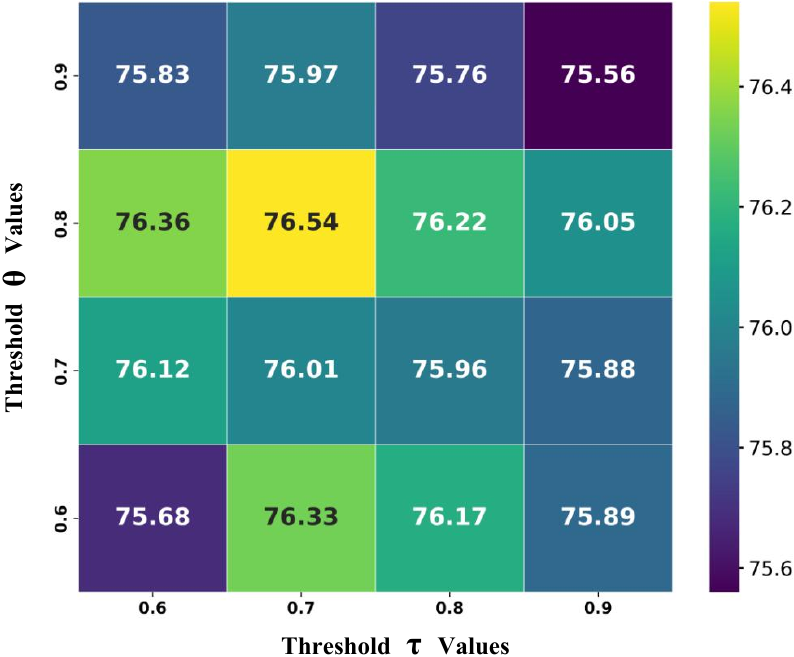}
	\caption{Ablation results of thresholds $\tau$ and $\theta$.}
    \label{fig:thres}
\end{wrapfigure}
(b) demonstrates our distribution-level matching approach, which improves coverage and precision by reducing false matches; (c) employs query region prototypes aggregation for better intra-object similarity, achieving higher foreground coverage and matching precision.

\noindent \textbf{Hyperparameter Evaluations.} Quantitative experiments are conducted to find a suitable number of agents $N_a$ and thresholds $\tau$ and $\theta$. As shown in Table~\ref{tab:agents_num}, increasing the number of agents improves performance by enabling finer and more comprehensive matching. However, too many agents introduce unnecessary and fragmented disturbances and reduce performance, with 100 agents found to be optimal. Fig.~\ref{fig:thres} illustrates the model's performance (mIoU) on the S3DIS dataset for different threshold settings, with optimal performance at $\tau = 0.7$ and $\theta = 0.8$,  respectively.

\section{Conclusion}
In this paper, we propose a novel Decoupled Localization and Expansion (DLE) approach to solve the \textbf{Incorrect background activation} and \textbf{Incomplete foreground mining} challenges in PC-FSS. Extensive experimental results on two standard benchmarks demonstrate that DLE performs favorably against state-of-the-art PC-FSS methods.


\section*{Acknowledgements}
This work was partially supported by the 
National Defense Basic Scientific Research Program (JCKY2021130B016), National Nature Science Foundation of China (Grant 62021001), and Youth Innovation Promotion Association CAS.

%
%
\bibliographystyle{splncs04}
\bibliography{main}
\end{document}